# Suspicious Object Recognition Method in Video Stream Based on Visual Attention

**Panqu Wang, Yan Zhang**
**Department of Electronic Engineering, Fudan University**

*Abstract*

We proposed a state-of-the-art method for intelligent object recognition and video surveillance based on human visual attention. Bottom-up and top-down attention are applied respectively in the process of acquiring interested object (saliency map) and object recognition. The revision of 4-channel PFT method is proposed for bottom-up attention and enhances the speed and accuracy. Inhibit of return (IOR) is applied in judging the sequence of saliency object "pop-out". Euclidean distance of color distribution, object center coordinates and speed are considered in judging whether the target is match and "suspicious." The extensive tests on videos and images show that our method in video analysis has high accuracy and fast speed compared with traditional method. The method can be applied into many fields such as video surveillance and security.

## I. Introduction and Motivation

Human vision has the ability of recognizing an interested object from a clustered and non-organized visual scene because of the existence of visual attention. Early psychological researchers have noticed the fact and started finding out the essence of attention since William James's proposal[1] that visual attention is like "having a focus, a margin and a fringe"[2] in 1890. In 1980 Treisman and Gelade proposed their famous Feature-Integration Theory of Attention [3]. They divided attention into two steps- feature segmentation and integration. Features like color, edge and orientation are extracted onto several maps; all feature maps are coded and integrated as a master "saliency map" on which only the most conspicuous area stands out. The saliency map contains information only about location. In 1995, Desimone and Duncan [4] perfected the attention theory by defining the attention into Bottom-up and Top-down process. The Bottom-up process separates figures from their background, while the top-down mechanisms select objects relevant to current behavior. We can regard the bottom-up process as "image-based" and top-down process as "task-independent" [5]. That means when we notice a natural scene, our visual system initially attends the pop-out object, for example, a human being stands on the grass. This bottom-up mechanism does not require us to know whether it is a human being or not, we just know the object is "relatively different" from its surroundings. Next, the top-down process received the knowledge of the object and started to analysis it by the prior knowledge of brain: is it a human? Do I know him? Usually the top-down result varies by the task given. The bottom-up and top-down process is in consistent of the function of dorsal stream, which is for spatial perception, and ventral stream, which is for object recognition [6] in our nerve center, so the theory has convincing biological basis.

In 1998, based on the feature-integration theory, L.Itii and Koch proposed the first computational model of visual attention-

Neuromorphic Vision C++ Toolkit (NVT) [7]. This method has high biological plausibility and acceptable processing speed, but severely dependent on the parameters selection. After NVT, several other computational methods are proposed: information maximization by Bruce and Tsotsos [8], discriminant saliency using center-surround method by Gao and Nuno [9], saliency using natural statistics (SUN) Bayesian method by Zhang et al[10]. All these methods consider the highly textured places are more salient, which is not usually accurate. In 2008, inspired by Hou's work that spectral residual (SR) by Fourier transform is a rapid approach to detect salient objects[11], Guo and Zhang proposed a bottom-up frequency domain method- PQFT in acquiring saliency map[12]. The method is significantly fast when compared with former methods. All approaches mentioned above are all bottom-up process. Actually tremendous work have been done in top-down attention based on each method respectively, such as SUN in top-down saliency by Kanan et al[13], and discriminant saliency used in object recognition by Dashan et al. in 2009[14].

In this work, we take video analysis into consideration. We want to detect the suspicious target (generally a human) in the video stream. For example, a person is wandering on the pavement for a long time; suddenly, he grabbed the bag of an old lady and started to run away. We hope the camera can detect the emergency, record the information of the suspicious person and make alert. The traditional methods, such as image subtraction, has fast speed however low accuracy, which is irresistible to tiny noise. Template matching, such as image convolution and correlation, process the image as a whole no matter how small and where the target is, so it is time consuming and impracticable in real-time video analysis. In fact, we do not even have a template to "learn" since the target can be a lot in the image and different as time goes by. In this work, we take visual attention into account. As human attention has the ability to notice the most salient object in a highly clustered scene and acquire the "saliency map" without any template and prior knowledge in bottom-up stage, we can apply this human intelligence into camera on video surveillance to make it time consuming and more precise. Considering video stream analysis or surveillance requires relatively fast speed (10-30 frames per second), we should adapt the fastest method in analyzing each frame to complete bottom-up and top-down attention process. More important, we should ensure the accuracy of interested area on saliency map and make the recognition process more efficient.

In this paper, we proposed the 4-channel phase Fourier transform (4PFT) based on PQFT method in acquiring saliency map in bottom-up process. Color (Red and Green, Yellow and Blue), intensity and motion channel are integrated to form the most salient area. After the saliency map completed, we start the Inhibit of Return (IOR) process in developing several most salient area and discarding other region of the image to advance the processing speed. In top-down stage, firstly we record the color distribution, center point, size and speed information of all salient area in successive frames, and then test whether they match any of the previous ones; if so, we update the location, size, color distribution and speed information of the target; if not, new area will be added in the memory. In addition, if the matched target has strange behavior- such as the sudden change of speed, we can notice the exception and mark the target and frame as "suspicious" for further identification. As human vision possesses the character that object looks small in the distance and big on the contrary, some parameters are modified in consistence with the biological fact. Finally, if the memory is full, we will delete some target which has not been seen for a long time, in accordance to human memory. The result shows our method really have excellent speed and high accuracy in identifying suspicious target in real-time video analysis.

The rest of this paper will be organized as follows: introduction to 4-channel PFT method will be presented; the IOR process and the definition of "suspicious" will be explained; procedure for object matching and suspicion justification will be shown; result of bottom-up method, measurement of top-down process, statistics based on extensive video and images analysis will be shown. At last, we will conclude our work and give some discussions.

## II. 4-Channel PFT Method in Acquiring Saliency Map

In 2007, Hou and Zhang [11] proposed that by extracting spectral residual (SR) of an image in spectral domain, we can construct the saliency map rapidly. Later, Guo et al. [15] showed that saliency map can also be calculated by the phase spectrum of an image's Fourier transform when set the amplitude spectrum to a nonzero constant value. However, this PFT method is only designed for gray-scale images and does not take other important vision factors into account (e.g., color pairs and motion). Due to the discovery of Stephen Engle et al. in 1997[16], red/green and blue/yellow color pairs are existed in human visual cortex. When neurons are excited by one color (e.g., green), they are inhibited by another color (e.g., red). Suppose the image frame captured at one time is F(t), t=1,2,…,n. R(t), G(t), B(t) and Y(t) are the red, green, blue and yellow color channel of the image. Therefore, the two color pair channels are designed as follows:

$$RG(t) = R(t) - G(t) \qquad (1)$$

$$BY(t) = B(t) - Y(t) \qquad (2)$$

Where the four broadly tuned color channels are defined as (3)-(6) [7]:

$$R(t) = r(t) - \frac{(g(t)+b(t))}{2} \qquad (3)$$

$$G(t) = g(t) - \frac{(r(t)+b(t))}{2} \qquad (4)$$

$$B(t) = b(t) - \frac{(r(t)+g(t))}{2} \qquad (5)$$

$$Y(t) = \frac{(r(t)+g(t))}{2} - \frac{|(r(t)-g(t))|}{2} - b(t) \qquad (6)$$

The intensity channel and motion channel are defined as (7) and (8).

$$I(t) = \frac{r(t)+g(t)+b(t)}{3} \qquad (7)$$

$$M(t) = |I(t) - I(t-\tau)| \qquad (8)$$

Where $\tau$ is the latency coefficient defined by our preferences.

Now we have built the four channels for the input image: two color pair channel, one intensity channel, and one motion channel. They are in consistent of the function of cells in primary visual cortex. Since all channels are almost independent, we can calculate each channel in PFT method, and integrate them to a unified saliency map. The PFT process is described below ($I(x, y)$ is defined as the input image).

$$f(u,v) = F(I(x,y)) \qquad (9)$$

$$p(u,v) = P(f(x,y)) \qquad (10)$$

$$sM(x,y) = \left\| F^{-1}[e^{i \bullet p(u,v)}] \right\|^2 \qquad (11)$$

$$O(x,y) = g(x,y) * (\lambda(RG) \cdot sM(RG) + \lambda(BY) \cdot sM(BY) + \lambda(I) \cdot sM(I) + \lambda(M) \cdot sM(M))$$

$$(12)$$

Where $F$ and $F^{-1}$ represent Fourier transform and Fourier inverse transform. $P(f)$ represents the phase spectrum of the input image. $sM(x, y)$ is the saliency map shown in x-y coordinate system. We should notice that we only use the phase spectrum while the amplitude spectrum is set to a nonzero constant. After we calculated all four channels using PFT method, we will receive four saliency maps. Then we integrate all of them by adding them with a weight $\lambda$ respectively. The weight can be set by our preference. In this work, motion in video analysis occupies much weight so we set $\lambda(M)$ a little higher. $g(x, y)$ is a disk filter where the radius is 3. Considering both accuracy and efficiency, we will resize the input image to the scale of 64*86 in proportion to the original scale.

We compared the 4-channel PFT method with PQFT method in a test of 100 natural images from a total database. The test results can be seen below:

**TABLE 1**
Average time cost in the test natural image

| Model    | 4PFT | PQFT  |
|----------|------|-------|
| Time(ms) | 16.8 | 136.9 |

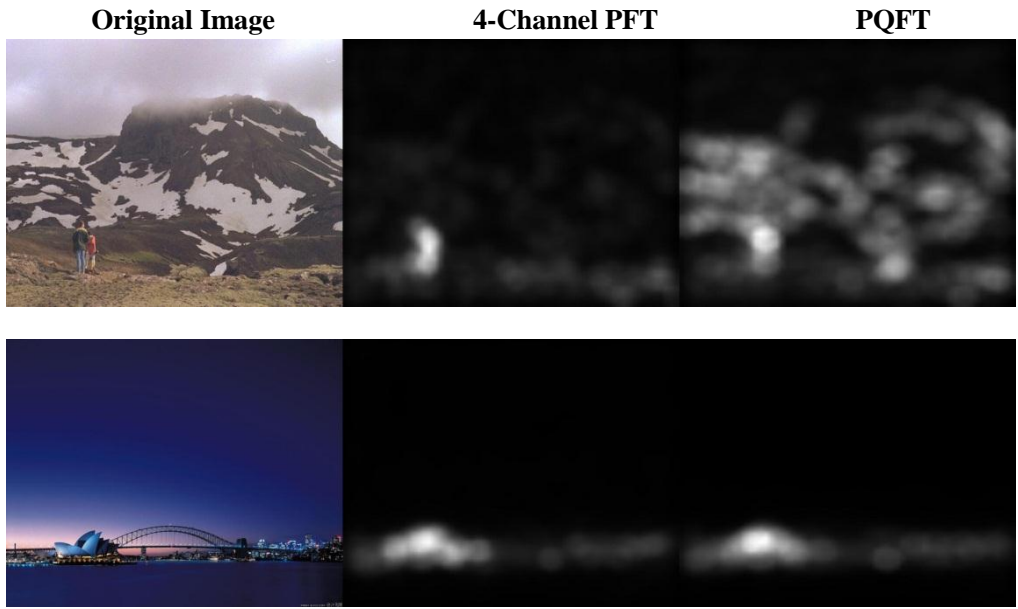

**Original Image**    **4-Channel PFT**    **PQFT**

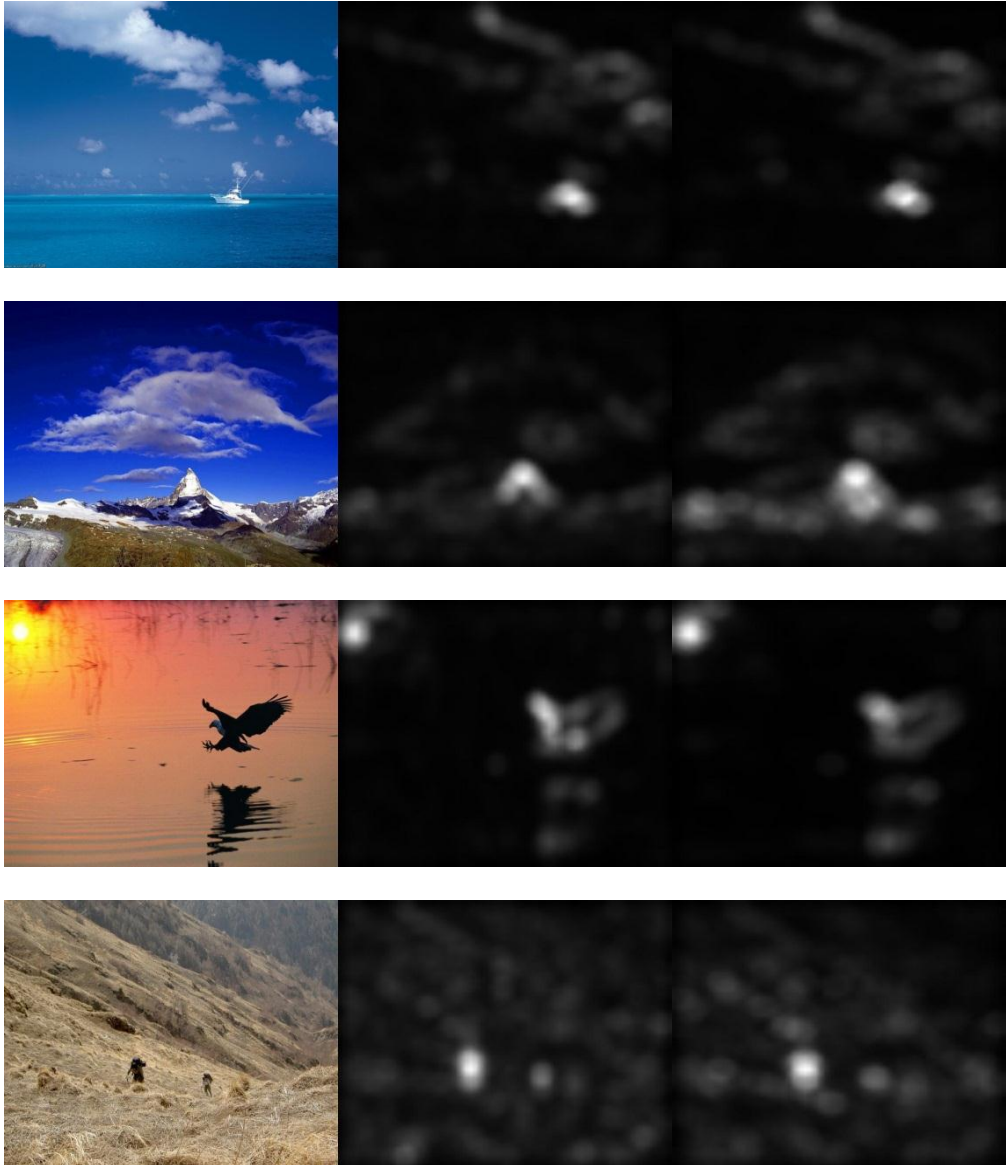

Fig.1.   Selected results calculated by 4-channel PFT and PQFT method. The images are available online at [17].

The result shows 4-Channel PFT has significant time advance compared with PQFT, which is in consistent of result in [12]. The output performance, however, is relatively undistinguishable. So the 4-Channel PFT method should be applicable in analyzing video stream. Here is the result of analyzing a video stream using 4-channel PFT and PQFT method respectively.

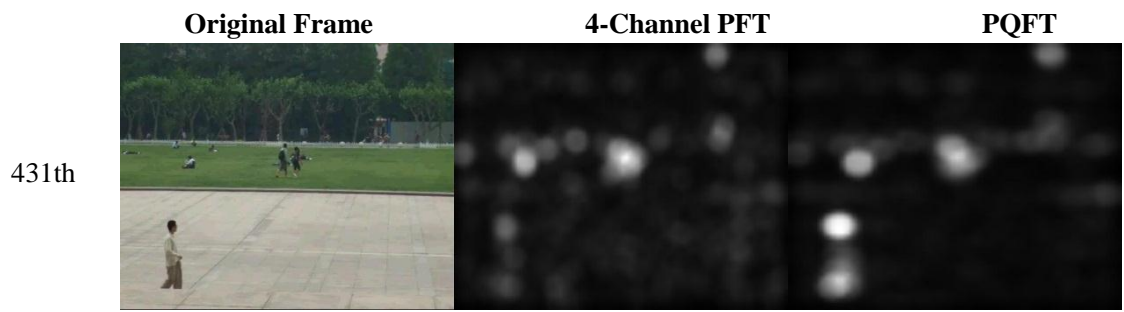

431th

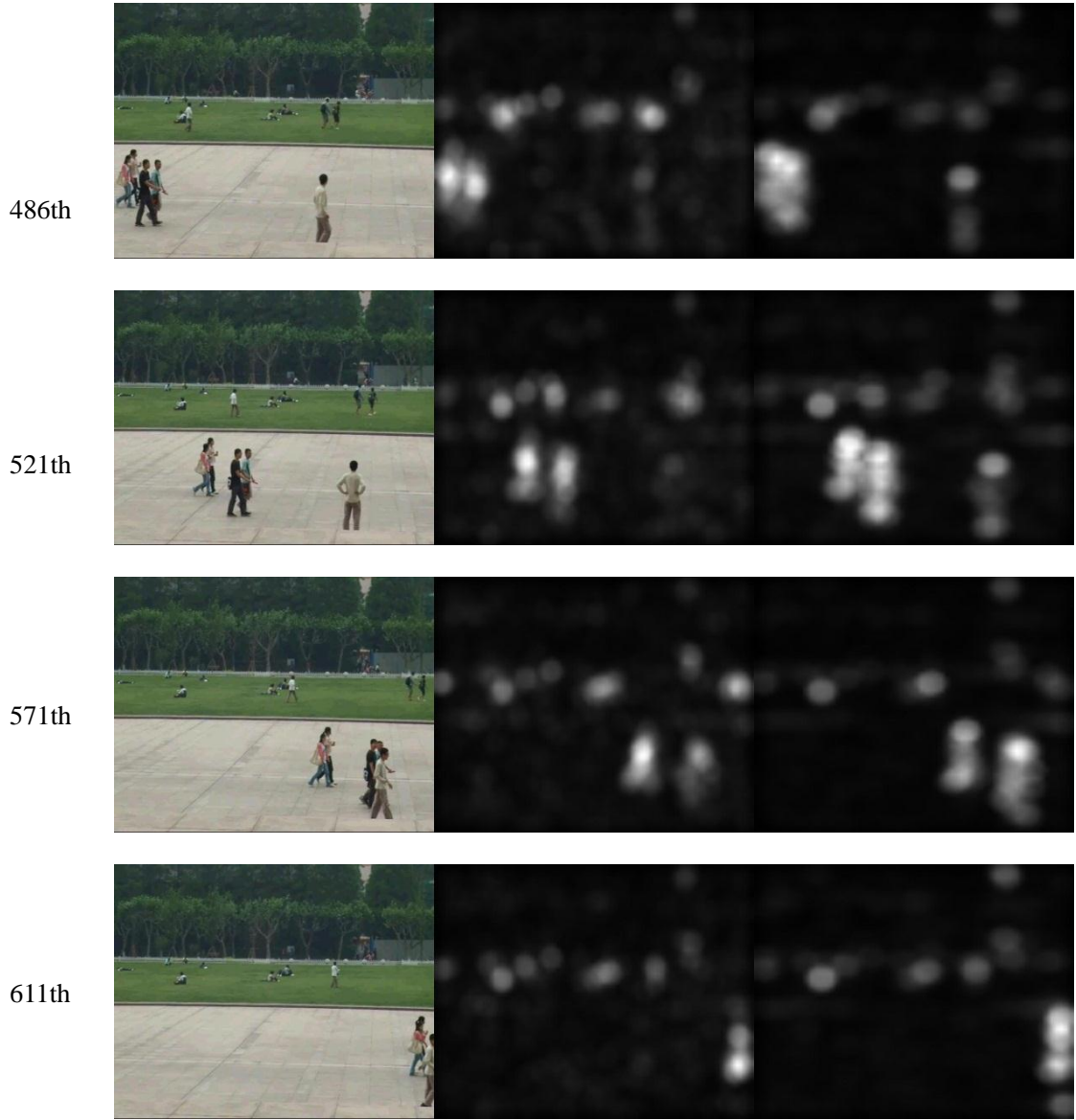

Fig.2. Selected results calculated by 4-channel PFT and PQFT method in a video stream.

The images are randomly selected in a series of video frames. We can see clearly that the interested target (wandering human) are successfully caught by the bottom-up attention. The accuracy and efficiency of 4-Channel PFT method provides stable foundation for the further top-down process of the video analysis.

# III. Top-down process in analyzing saliency map

## A. Inhibition of Return (IOR) process in acquiring interested area

After we get the saliency map from bottom-up attention, it is time for us to extract the interested objects subsequently. According to the early visual research theory discovered by Posner and Cohen [18], when the most salient area is noticed at the first sight, by following a shift of attention away from the cued area, targets at that location are handled less efficiency than at other places. When the inhibition of return (IOR) process is applied onto the saliency map, we can conclude that the most salient area, which has the highest intensity of the image, should be noticed first. Then it is suppressed, letting the visual system search for the next salient area. This process is not finished until all residual in the saliency map cannot catch attention.

In our model, this IOR process can be defined as follows:

Suppose we get a saliency map $sM(t)$ which originated from original frame $F(t)$ at time $t$. The $n^{th}$ search starts by finding the most salient (highest intensity value) point in $sM_n(t)$. We want to find the $n^{th}$ object candidate area (OCA) by finding the zone where the intensity value $O(x, y)$ of all points in the area

$$\alpha \cdot O_i^{\max} \leq O(x, y) \leq O_i^{\max} \qquad (13)$$

Where $\alpha$ is the user-defined threshold which can affect the area's size. We can see the result by setting different values of $\alpha$:

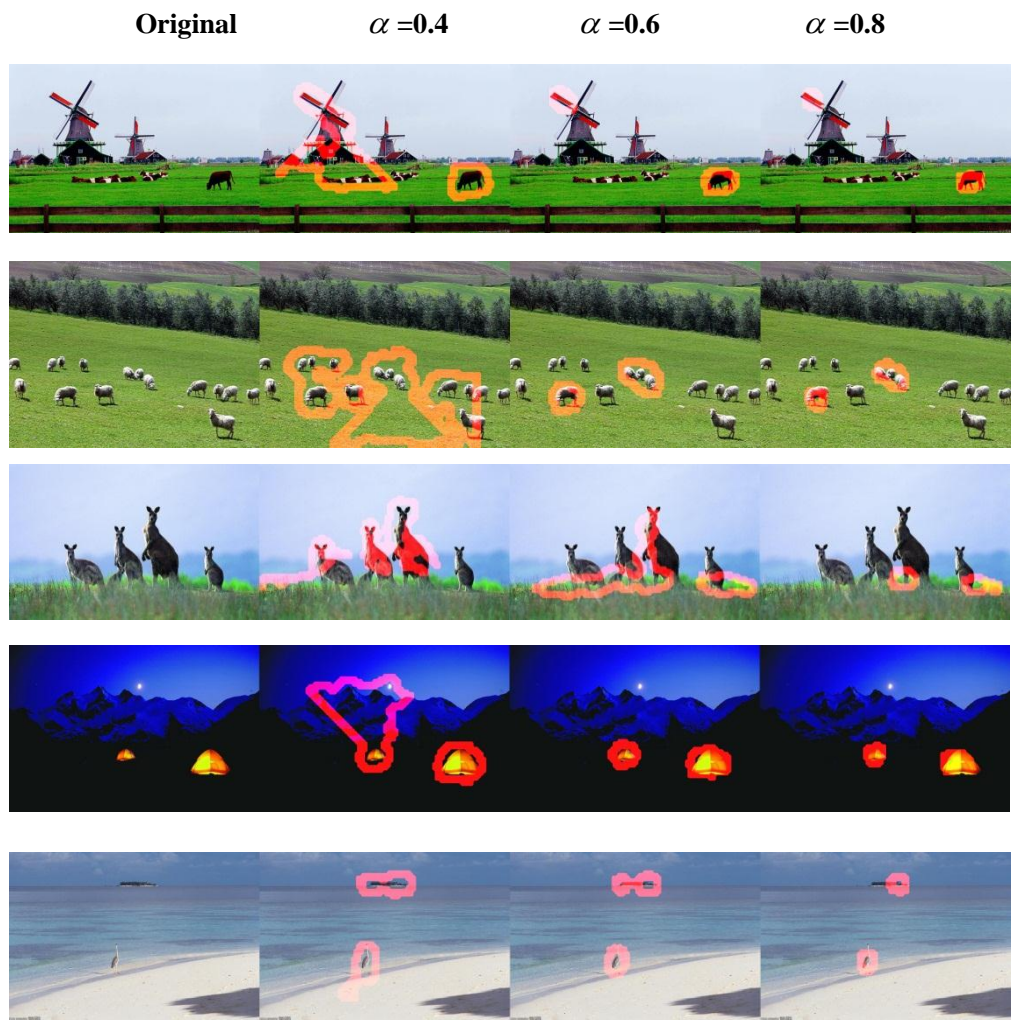

Fig.3. Selected results calculated by 4-channel PFT followed by IOR process with different $\alpha$. The images are available online at [17].

The OCA area is circled by red rings. We can see the area size varies a lot by the change of $\alpha$. When the $\alpha$ increases from 0.4 to 0.8, the threshold becomes bigger, so the area becomes smaller. This essence of IOR process is to find the requested 8-connected neighborhood region of $(x_i, y_i)$ continuously. After extensive tests on our videos, we get the value of $\alpha = 0.55$ to reach the optimal result.

When $OCA_n$ is found, the corresponding area of the saliency map should be suppressed to zero (black) and the whole saliency map becomes $sM_{n+1}$. Then a new IOR process starts. The total number $N$ is set by the user. The different selection of value $N$ will influence the output effectiveness. The results are as follows: ($\alpha$ =0.55)

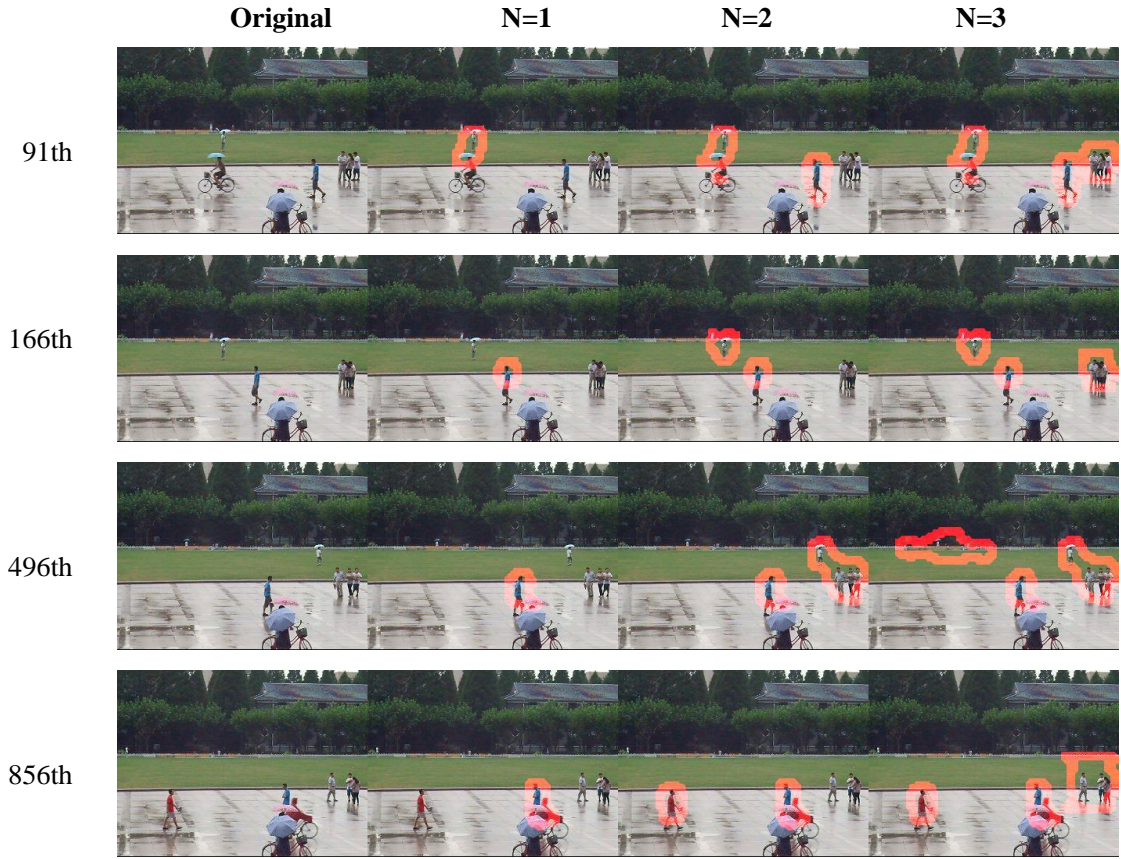

Fig.4. Selected results calculated by 4-channel PFT followed by IOR process with different N in a test video.

In our paper, $N$ is set as 4 to ensure that the most salient area and suspicious target can be contained in the areas.

There is one fact we should notice. According to our basic knowledge for human vision, object looks small in the distance and big on the contrary. So the threshold of $\alpha$ should vary according to the distance. When reflected onto the image, the upper rows represent "far" and the lower rows represent "near". So the threshold $\alpha$ will be set larger if the distance is far and vice versa. This is because the object far away itself is blurry to human vision, so there is no necessity to identify them so delicate. Imagine a person walked near from far away, he will be looked larger and larger when he approaches. So the OCA area should grow to cover the whole object. That is why $\alpha$ is smaller when the distance is near.

Another fact we should notice is the size of OCA area. When N grows bigger, as the most salient area has been moved away, the remainder area perhaps has the same intensity level. That means, when we continue the IOR process, as $\alpha$ is fixed, the surrounding area with suitable intensity level may be extensive- even the rest of the image. However, this kind of area is usually useless- they are less important (because less salient) and usually not the target because the area is so large. By our experience, the suspicious people in the test video of size 64*80 should have the size less than 200 pixels. So we decide to discard the blocks whose size is bigger than 300 pixels to cut off the time cost and made the whole process more likely to the reality.

*B. How to detect the "suspicious" target in video frames?*

After the Inhibition of Return process ends, we will receive $N$ salient area of each frame. In preparation for the object recognition process, we should record the following information of each salient area: size, color distribution, center location, speed and time it appears. All of the information are recorded in a struct $S$:

S= repmat(struct('time',zeros(1,2), 'size', zeros(1,4),'color', zeros(1,4,11),'position', zeros(2,2),'speed',zeros(1,1)), [1 num]);

(14)

In function (14), 'size' records the size of 4 salient areas in one frame; 'color' is for the color distribution of one area; 'position' records the center point of corresponding salient area; 'speed' shows the speed of current target; 'time' is the reference of judging whether the target is suspicious and deciding which target to delete if the memory is full; num is the maxim capacity of the struct S.

In our model, the simplified definition of suspicious target is: Owing to the prerequisite of object recognition, the object will be considered suspicious when its velocity has sudden change. Of course we can define more complicated principles in the model for "suspicious", no matter it is sudden-based or accumulation- based just by adding additional variables.

The first step to identify whether one object is suspicious or not is to judge if it is a target appeared before. In another word, object recognition should be completed before determining the suspicion. The suspicion without prior knowledge is nonsense just by bottom-up information. The steps of object recognition are defined as follows:

*a. Color match*
Color distribution is the most direct information acquired from the salient area. In the IOR process, we would have recorded the color information of each pixel of one of the N salient area. The color information contains the red, green, blue and intensity values, which is defined the same as (7). After we get the four-channel color information, we make the histogram distribution of each channel. We make ten intervals between 0, the darkest and 1, the brightest intensity level. We count the possibility distribution of each interval. Obviously, all possibilities in the 10 intervals must be added as 1. After the four-channel distribution come out, we calculate the Euclidean distance between current salient area with each previous saved salient area. We also made the threshold $\eta$ for the Euclidean distance. The measurement of color match can be estimated as follow:

$$ColorMatch = 1 - \frac{D(S_i, S_j)}{\eta} \quad (15)$$

Where $D(S_i, S_j)$ stands for the Euclidean distance between current area $S_i$ and any previous area, which is saved in struct S, marked as $S_j$. Usually

$$D(S_i, S_j) = \sum_{k=1}^{4}(\sum_{l=1}^{10}(h_{li} - h_{lj}))C_k \quad (16)$$

In this equation, $C_k$ ($k=1,2,3,4$) is for the four color channels; $h_l$ represent the histogram possibility distribution of each interval from 1 to 10 of each color channel. The output *ColorMatch* must be a value between 0 to 1. If $D(S_i, S_j)$ is greater than $\eta$, the output should be set to zero because they are not match at all. Experience showed that $\eta$ is typically set as 0.6.

*b. Position match*

Besides the color distribution of the salient area, position is another factor to take into account. We cannot imagine the object is in the upper left corner of the camera can reach the lower right corner, even if they have similar color distribution, because nothing can move so fast in just 0.05s (Suppose the camera speed is 20 frames per second). The process of position match is comparably simple. We initially record the center of current saliency area. The horizontal location of the center is determined by the upmost and bottom x-coordinate obtained in IOR process, while the vertical location is determined by the leftmost and rightmost y-coordinate. The distance is measured by using the following function:

$$\overrightarrow{D_d}(x_{s_{ij}}, y_{s_{ij}}) = 2 \cdot (x_j, y_j)^T - (x_k, y_k)^T - (x_i, y_i)^T$$
$$= [(x_j, y_j)^T - (x_i, y_i)^T] - [(x_k, y_k)^T - (x_j, y_j)^T] \quad (17)$$

The $(x_i, y_i)^T$ is the current salient area's center point, $(x_j, y_j)$ is the $j^{th}$ salient area's center which has been saved in struct S, while $(x_k, y_k)^T$ is the previously matched area's center location compared with $S_j$. This area is marked as $S_k$. We use 3 areas in the position match to reduce the error of distance measurements. We can conclude that if the target is in uniform rectilinear motion, the result of $\overrightarrow{D_d}(S_i, S_j)$ will be zero. Also a threshold $\mu$ for position match is set and the result is described below:

$$PositionMatch = 1 - \frac{\left\|\overrightarrow{D_d}(x_{s_{ij}}, y_{s_{ij}})\right\|^2}{\mu}$$
$$(18)$$

Where

$$\left\|\overrightarrow{D_d}(x_{s_{ij}}, y_{s_{ij}})\right\| = \sqrt{(x_{s_{ij}})^2 + (y_{s_{ij}})^2} \quad (19)$$

When the object is in uniform rectilinear motion, *PositionMatch* will be 1. The *PositionMatch* will decrease otherwise.

Usually $\mu$ should vary by the video's frame rate, size of the saliency map and interval between frames. Also in accordance to the rule that object looks small in the distance and big on the contrary, $\mu$ should be set larger if the target is close to the bottom of the frame and vice versa because our vision thought the close object moves faster than the remote ones.

*c. Update and delete*

The *ColorMatch* and *PositionMatch* are considered comprehensively in judging whether the object is similar to any of previous ones. In our model, we take the color information into more consideration because each object's color identity can be rather distinguished. The final decision is:

$$Decision = 0.7 * ColorMatch + 0.3 * PositionMatch \quad (20)$$

After extensive test on the matched area, the *Decision* usually have the value higher than 0.7, even exceeds 0.9 if the two objects are still frame by frame. So when we decide to update one object, we should update the information of color distribution, size and the position at the same time.

However, when there are so many new objects added into the memory (struct S, number of the memory contains will be set by variable num, here set to 1000), they will finally exceed the capacity. Like our brains forget things we have long not been used or remembered, the model should "forget" the salient areas long not have been updated. So when the number of interested area exceeds the maximum, we will delete the area appeared least and not have been updated for a long time:

$$DelDecsion = 0.8*DelCount - 0.2*DelInterval \qquad (21)$$

The *DelCount* is the number of each area appears, while *DelInterval* is the time interval between the current frame and the latest frame one area appears. If the *DelCount* is small and *DelInterval* is great, the *DelDecsion* will be little. We sort all areas contained in the struct and delete the area with least value of *DelDecsion* ,then add new area to that place.

After all the object recognition tasks complete, the justification of suspicious target will be simplified. The "suspicious" is defined here as the sudden change of the matched target's speed. Imagine a wandering people find the victims, step forward, rob something from him and run away immediately- our camera should notice his run-away action and lock the suspicious target. Actually we do this job simultaneously with the update process. When the position updates, we can measure the speed of the object by counting the distance of center point between two salient areas. When the speed change exceeds one threshold $\varepsilon$, we should regard the target as suspicious:

$$\left|\frac{dV_i}{dt}\right| > \varepsilon \qquad (22)$$

Where $dV_i$ is the change of velocity. $\varepsilon$ also should vary by distance. The nearby threshold should be larger while the faraway threshold should be smaller.

## IV. Experimental Results

To evaluate the performance of our video analysis model based on visual attention, we should test the program under different circumstances: different angle, changeable weather, various locations and diverse scene. Finally we test our attention model on five different videos.

Note: The test machine is Intel(R) Core(TM)2 Duo CPU T7100@ 1.80GHz 1.79GH,1.96GB memory. Test environment: Matlab R2010a. Information of the five test video is as follows:

**Table 2 Information of the 5 test videos**

| Number | Frames | Targets |
|--------|--------|---------|
| Video1 | 210    | 440     |
| Video2 | 228    | 639     |
| Video3 | 170    | 366     |
| Video4 | 117    | 300     |
| Video5 | 359    | 295     |

The most intuitive results of finding suspicious target are showed below:

| Video1 | Video2 | Video3 |

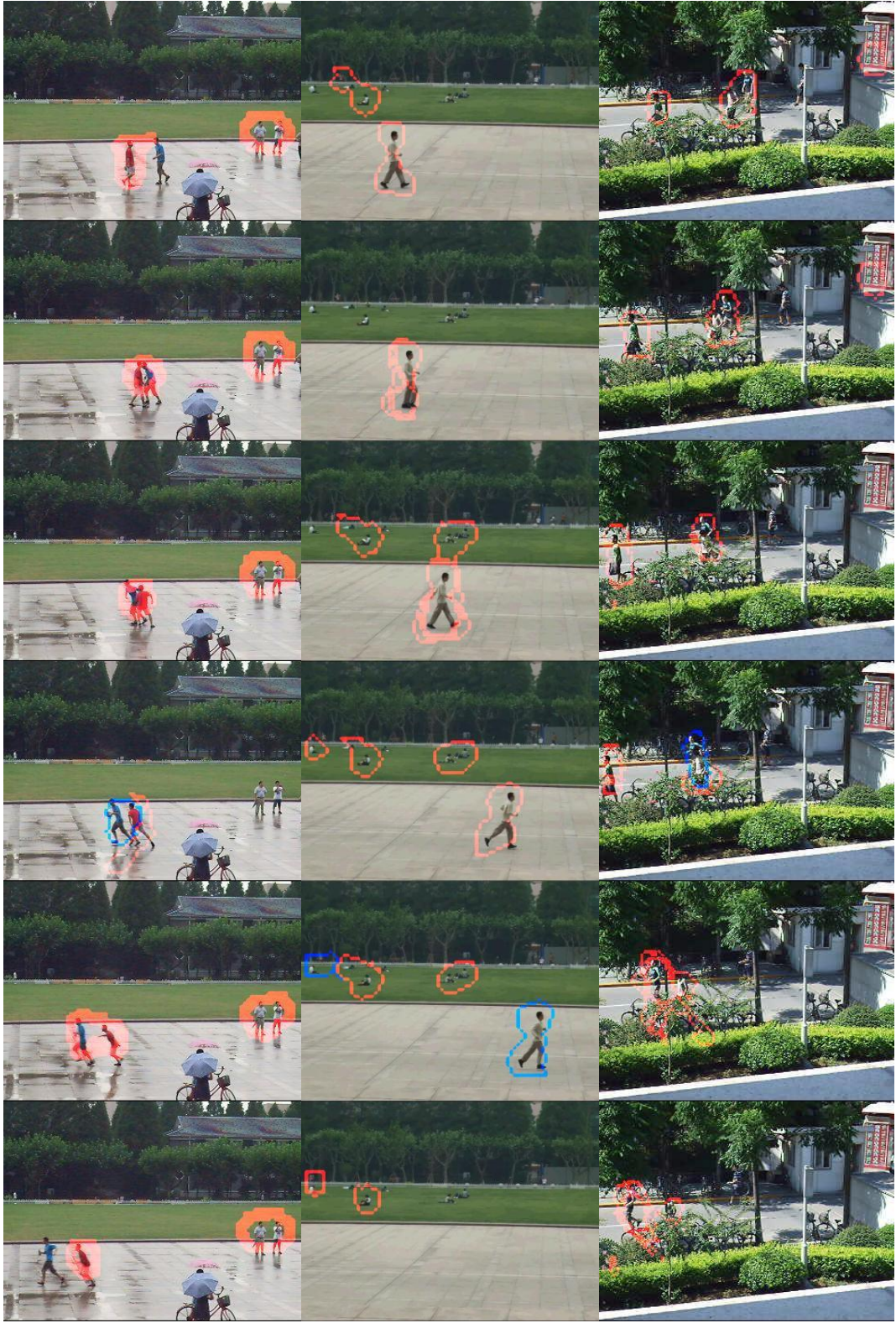

Fig.5. Selected results of the whole visual attention model applied on different video steams. The salient area is circled by red while the suspicious target is circled by blue.

We can see the three different scenes. Video 1 record a man approaches the victim, grabbed his umbrella and started to run away. Our model records the moment the target is ready to escape clearly by locking the target by the blue ring. Video2 simply shows a walking person starts to run. Our model detects the running time in the $4^{th}$ frame from top. It verifies our model is really sensitive to speed change. Video 3 is a highly clustered scene. A people circled by red in the former frames starts to run—although he does not attack anybody. We just detect the exceptional action according to our rules. It will be helpful if the safety guard want to identify the suspicious people by simply recalling the frames contain blue circles.

*A. Executing time*

The executing time is the one of the most important take into consideration according to our model. If the speed is not fast enough to deal with normal video streams, the model would be useless at all. The test on 5 different videos shows the time cost as follows:

**Table 3 Executing time of our attention model**

| Number | Bottom-up(ms) | Overall(ms) |
|---|---|---|
| Video1 | 21.45 | 93.76 |
| Video2 | 21.34 | 106.01 |
| Video3 | 19.23 | 111.20 |
| Video4 | 22.01 | 67.96 |
| Video5 | 27.14 | 126.48 |
| Average | 22.23 | 101.08 |

We can see from the Table 2 that the overall executing time (contain both Bottom-up and Top-down attention) for one frame is approximately 0.1s in average. The result is derived when $N$ (number of interested salient area per frame) equals to 4, which in general exceeds the request. So the overall time should be lower if we change $N$. Thus, we can meet with the basic requirement of video processing (10 frames per second). If we transform the Matlab code to C, the speed should be much faster.

*B. Size of salient area*

One advantage using visual attention is because we just notice several most salient areas in one big scene, which save many time compared with whole-image processing. This advantage in catching the most important facts while discarding other worthless is the key factor of our attention model. The average size of each of the four areas is as follows:

**Table 4 Average salient area size (pixels) of each video**

| No. | 1 | 2 | 3 | 4 |
|---|---|---|---|---|
| Video1 | 119.0 | 130.9 | 148.1 | 141.2 |
| Video2 | 106.1 | 96.0 | 96.2 | 73.2 |
| Video3 | 107.8 | 97.3 | 84.1 | 61.8 |
| Video4 | 85.1 | 96.0 | 109.8 | 120.7 |
| Video5 | 121.8 | 147.2 | 133.7 | 129.5 |
| Average | 108.0 | 113.5 | 114.4 | 109.3 |

Total average = 445.2pixels

$$\text{Total percentage} = \frac{Total average}{FrameSize(64*86)}$$
$$= 8.1\%$$

(20)

Since we only process 8.1% of the whole image, the enhancement of speed is easy to be understood. Please note we discard the size above 300 pixels, or the rate should be higher.

*C. Match score and false-alert rate*

The most important measurement in top-down process is the match score, which determines whether the performance of our model is good. In the five test videos, in order to obtain the accurate match score, we should analyze the video frame by frame. If the object is match between current frame and any other previous one correctly, we marked it as "true". If the match result is wrong, (i.e. match the current object with previous one which is not the object at all), we marked it as "false". After we analyze the whole video, we can calculate

the match performance by

$$Score = \frac{N(true)}{N(true) + N(false)} \quad (24)$$

The match rate measured in the five videos is:

**Table 5 Match score of the attention model**

| Number | Overall | False | Score |
|---|---|---|---|
| Video1 | 440 | 10 | 97.72% |
| Video2 | 639 | 18 | 97.18% |
| Video3 | 366 | 13 | 96.45% |
| Video4 | 300 | 10 | 96.67% |
| Video5 | 295 | 9 | 96.27% |
| Average Score: 96.86% | | | |

We can wee from above that the average match score of our model test on videos is stable (around 97%) and relatively high to meet with object recognition requirements.

Another measurement is called false-alert rate. After we checked the suspicious target, we found 100% of them are captured by the camera using our attention models. However, this causes another problem. The more area are circled and thought suspicious, the accurate rate will undoubtedly increase. The most extreme condition is that we consider all points of the frame is suspicious—and the score must be 100 because all suspicious area must have been circled. However, it is often impracticable because a lot of non-suspicious target is circled too. We define this kind of acquisition- considered suspicion when it is not- as false alert. This kind of attention should be avoided because it may confuse our judgments. Typically the false-alert rate arises when the suspicion-identification rate is high. The function for calculating false-alert rate is:

$$FalseAlert = \frac{N(false)}{N(suspicious)} \quad (25)$$

Where $N(false)$ is the false-alert counter while $N(suspicious)$ is the total number of considered suspicious.

Here is the false-alert rate of our model:

**Table 6 False-alert rate of the attention model**

| Number | Suspicious | False | rate |
|---|---|---|---|
| Video1 | 36 | 6 | 16.70% |
| Video2 | 75 | 15 | 20.00% |
| Video3 | 238 | 42 | 11.20% |
| Video4 | 67 | 12 | 17.90% |
| Video5 | 64 | 10 | 15.60% |
| Average Rate: 16.28% | | | |

To reduce the false-alert rate, we should develop more advanced method in recognizing suspicious target. At the same time, we should reach the balance between the match score and false-alert rate to obtain optimal performance of our models.

# V. Conclusions and discussions

In conclusion, our work can be described as developing a frequency-domain bottom-up attention method- 4-Channel PFT in acquiring saliency maps from video sequence, then applying IOR process to obtain interested area for object recognition, finally searching for the suspicious target in the video stream. The bottom-up and top-down attention is used respectively in finding salient areas and object recognition. The bottom-up 4-channel PFT method has speed advantage and good performance in acquiring saliency map both in natural images and movies. The IOR process follows the biological mechanisms to notice different interested areas. We can also set the number of wanted areas by modifying $N$. The top-down task-orientated process considers color distribution, area size, center point and speed to determine the match scores. We will update the matched object and delete the most sacred ones if the memory is full. The result shows our model has excellent executing time in analyzing each frame in the video sequence. The match score is high and the false-alert rate is acceptable.

We believe our work has great potentials in applying to many other engineering fields such as video surveillance and safety guard. It is necessary to test the model into many other complicated scenes. The principle of defining suspicious can be more delicate, as the model has high potential to be extended. Our model should compare with other top-down models to acquire more convincing statistics, such the ROC curve between match scores and false-alert rate. How to decrease the pseudo scene rate without eliminating the match score is another interested topic to be discussed.

**Refrences:**

[1] W. James, The Principles of Psychology. New York: Holt, 1890.
[2] Eriksen C & Hoffman J. (1972). Temporal and spatial characteristics of selective encoding from visual displays. Perception & Psychophysics, 12(2B), 201–204.
[3] Treisman, A., & Gelade, G. (1980). A feature-integration theory of attention. Cognitive Psychology, 12, 97-136.
[4] R.Desimone and J.Duncan.,1995. Neural mechanisms of selective visual attention. Annu. Rev. Neurosci. 1995. 18: 193-222
[5] Itti, L., & Koch, C. (2001). Computational modelling of visual attention. Nature Reviews, Neuroscience, 2, 194–203.
[6] Ungerleider, L.G., Mishkin, M., 1982. Two cortical visual systems.In: Ingle, D.J., Goodale, M.A., Mansfield, R.J.W. (Eds.), Analysis of Visual Behavior. InMIT Press, Cambridge, pp. 549–586.
[7] L. Itti, C.Koch, and E. Niebur, "A model of saliency-based visual attention for rapid scene analysis," IEEE Trans. Pattern Anal. Mach. Intell., vol. 20, no. 11, pp. 1254–1259, Nov. 1998.
[8] Bruce, N., & Tsotsos, J. (2006). Saliency based on information maximization. In Y. Weiss, B. Scho¨lkopf, & J. Platt (Eds.), Advances in neural information processing systems 18 (pp. 155–162). Cambridge, MA: MIT Press.
[9] Itti, L., & Koch, C. (2001). Computational modelling of visual attention. Nature Reviews, Neuroscience, 2, 194–203.
[10] Zhang, L., Tong, M. H., Marks, T. K., Shan, H., & Cottrell, G. W. (2008). SUN: A Bayesian framework for saliency using natural statistics. Journal of Vision, 8(7), Pt. 32, 1-20.
[11] X. Hou and L. Zhang, "Saliency detection: A spectral residual approach," presented at the CVPR, 2007.
[12] C. Guo, L. Zhang, "A Novel Multiresolution Spatiotemporal Saliency Detection Model and Its Applications in Image and Video Compression," IEEE Trans. Image Processing, 2010. 19(1): 185-198.
[13] C.Kanan, Tong, M. H., Zhang, L., & Cottrell, G. W. (2009). SUN:Top-down saliency using natural statistics. Visual Cognition, Vol.17, Issue 6&7, 979 - 1003
[14] D.Gao., S.Han., & N. Vasconcelos.(2009). Discriminant Saliency, the Detection of Suspicious Coincidences, and Applications to Visual Recognition. IEEE Trans. Pattern Analysis and Machine Intelligence, Vol.31, Issue 6, 989-1005.
[15] C. L. Guo, Q. Ma, and L. M. Zhang, "Spatio-temporal saliency detection using phase spectrum of quaternion Fourier transform," presented at the CVPR, 2008.
[16] S. Engel, X. Zhang, and B. Wandell, "Colour tuning in human visual cortex measured with functional magnetic resonance imaging," Nature, vol. 388, no. 6,637, pp. 68–71, Jul. 1997.
[17] [Online]. Available: http://homepage.fudan.edu.cn/~clguo
[18] M. I. Posner and Y. Cohen, "Components of visual orienting," in Attention and

Performance, H. Bouma and D. G. Bouwhuis, Eds. Hillsdale, NJ: Erlbaum, 1984, vol. 10, pp. 531–556.